\documentclass[10pt, a4paper]{article}
\usepackage{lrec2016}
\usepackage{multibib}
\newcites{languageresource}{Language Resources}
\usepackage{times}
\usepackage{url}
\usepackage{latexsym}
\usepackage{enumitem}
\usepackage{amsmath}
\usepackage{xspace}
\usepackage{graphicx}
\usepackage{arydshln}
\usepackage{hyperref}
\usepackage{ragged2e}
\usepackage{soul}

\usepackage{xcolor}
\hypersetup{
    colorlinks,
    linkcolor={red!20!black},
    citecolor={blue!20!black},
    urlcolor={blue!15!black}
}

\graphicspath{ {fig/} }

\usepackage[none]{hyphenat}

\newcommand{\rouge}{\textsc{Rouge}\xspace}
\newcommand{\rg}{\textsc{Rouge}}

\newcommand{\ser}{\textsc{Sera}\xspace}
\newcommand{\sr}{\textsc{Sera}}

\usepackage[splitrule]{footmisc}

\usepackage{epstopdf}
\usepackage[latin1]{inputenc}

\title{Revisiting Summarization Evaluation for Scientific Articles}

\name{Arman Cohan and Nazli Goharian}

\address{Information Retrieval Lab \\
         Department of Computer Science \\
         Georgetown University \\
         arman@ir.cs.georgetown.edu, nazli@ir.cs.georgetown.edu\\}

\abstract{
Evaluation of text summarization approaches have been mostly based on metrics that measure similarities of system generated summaries with a set of human written gold-standard summaries. The most widely used metric in summarization evaluation has been the \rouge family. \rouge solely relies on lexical overlaps between the terms and phrases in the sentences; therefore, in cases of terminology variations and paraphrasing, \rouge is not as effective. Scientific article summarization is one such case that is different from general domain summarization (e.g. newswire data). We provide an extensive analysis of \rg's effectiveness as an evaluation metric for scientific summarization; we show that, contrary to the common belief, \rouge is not much reliable in evaluating scientific summaries. We furthermore show how different variants of \rouge result in very different correlations with the manual Pyramid scores. Finally, we propose an alternative metric for summarization evaluation which is based on the content relevance between a system generated summary and the corresponding human written summaries. We call our metric \ser (Summarization Evaluation by Relevance Analysis). Unlike \rouge, \ser consistently achieves high correlations with manual scores which shows its effectiveness in evaluation of scientific article summarization. \\ 
\newline 
\vspace{16pt}
\Keywords{Summarization, Evaluation, Scientific articles} }

\begin{document}

\maketitleabstract

\section{Introduction}

Automatic text summarization has been an active research area in natural language processing for several decades. To compare and evaluate the performance of different summarization systems, the most intuitive approach is assessing the quality of the summaries by human evaluators. However, manual evaluation is expensive and the obtained results are subjective and difficult to reproduce \cite{giannakopoulos2013summary}. To address these problems, automatic evaluation measures for summarization have been proposed. \textsc{Rouge} \cite{lin2004rouge} is one of the first and most widely used metrics in summarization evaluation. It facilitates evaluation of system generated summaries by comparing them to a set of human written gold-standard summaries. It is inspired by the success of a similar metric \textsc{Bleu} \cite{papineni2002bleu} which is being used in Machine Translation (MT) evaluation. The main success of \textsc{Rouge} is due to its high correlation with human assessment scores on standard benchmarks \cite{lin2004rouge}. \rouge has been used as one of the main evaluation metrics in later summarization benchmarks such as TAC\footnotemark[1] \cite{owczarzak2011overview}.

\footnotetext[1]{Text Analysis Conference (TAC) is a series of workshops for evaluating research in Natural Language Processing}



Since the establishment of \rouge, almost all research in text summarization have used this metric as the main means for evaluating the quality of the proposed approaches. The public availability of \rouge as a toolkit for summarization evaluation has contributed to its wide usage. While \rouge has originally shown good correlations with human assessments, the study of its effectiveness was only limited to a few benchmarks on news summarization data (DUC\footnotemark[2] 2001-2003 benchmarks). Since 2003, summarization has grown to much further domains and genres such as scientific documents, social media and question answering. While there is not enough compelling evidence about the effectiveness of \rouge on these other summarization tasks, published research is almost always evaluated by \rouge. In addition, \rouge has a large number of possible variants and the published research often (arbitrarily) reports only a few of these variants.

\footnotetext[2]{Document Understanding Conference (DUC) was one of NIST workshops that provided infrastructure for evaluation of text summarization methodologies (\url{http://duc.nist.gov/}).}

By definition, \rouge solely relies on lexical overlaps (such as n-gram and sequence overlaps) between the system generated and human written gold-standard summaries. Higher lexical overlaps between the two show that the system generated summary is of higher quality. Therefore, in cases of terminology nuances and paraphrasing, \rouge is not accurate in estimating the quality of the summary.

We study the effectiveness of \rouge for evaluating scientific summarization. Scientific summarization targets much more technical and focused domains in which the goal is providing summaries for scientific articles. Scientific articles are much different than news articles in elements such as length, complexity and structure. Thus, effective summarization approaches usually have much higher compression rate, terminology variations and paraphrasing \cite{teufel2002summarizing}.

Scientific summarization has attracted more attention recently (examples include works by \newcite{abu2011coherent}, \newcite{qazvinian2013generating}, and \newcite{cohan2015scientific}). Thus, it is important to study the validity of existing methodologies applied to the evaluation of news article summarization for this task. In particular, we raise the important question of how effective is \textsc{Rouge}, as an evaluation metric for scientific summarization? We answer this question by comparing \rouge scores with semi-manual evaluation score (Pyramid) in TAC 2014 scientific summarization dataset\footnotemark[1]. Results reveal that, contrary to the common belief, correlations between \textsc{Rouge} and the Pyramid scores are weak, which challenges its effectiveness for scientific summarization. Furthermore, we show a large variance of correlations between different \rouge variants and the manual evaluations which further makes the reliability of \rouge for evaluating scientific summaries less clear. We then propose an evaluation metric based on relevance analysis of summaries which aims to overcome the limitation of high lexical dependence in \textsc{Rouge}. We call our metric \ser (\textit{Summarization Evaluation by Relevance Analysis}). Results show that the proposed metric achieves higher and more consistent correlations with semi-manual assessment scores. 

\footnotetext[1]{\url{http://www.nist.gov/tac/2014/BiomedSumm/}}

Our contributions are as follows: 
\begin{itemize}[leftmargin=8pt,label={--}]
\item Study the validity of \textsc{Rouge} as the most widely-used summarization evaluation metric in the context of scientific summarization.
\item Compare and contrast the performance of all variants of \rouge in scientific summarization.
\item Propose an alternative content relevance based evaluation metric for assessing the content quality of the summaries (\ser).
\item Provide human Pyramid annotations for summaries in TAC 2014 scientific summarization dataset.\footnotemark[2]
\end{itemize}

\footnotetext[2]{\RaggedRight The annotations can be accessed via the following repository: \url{https://github.com/acohan/TAC-pyramid-Annotations/}}

\section{Summarization evaluation by \rouge}
\rouge has been the most widely used family of metrics in summarization evaluation. In the following, we briefly describe the different variants of \rouge:

\begin{itemize}[leftmargin=8pt,label={--}]
\setlength\itemsep{2pt}
\item \rg-N:
\rg-N was originally a recall oriented metric that considered N-gram recall between a system generated summary and the corresponding gold human summaries. In later versions, in addition to the recall, precision was also considered in \rg-N, which is the precision of N-grams in the system generated summary with respect to the gold human summary. To combine both precision and recall, F1 scores are often reported. Common values of N range from 1 to 4.

\item \rg-L:
This variant of \rouge compares the system generated summary and the human generated summary based on the Longest Common Subsequences (LCS) between them. The premise is that, longer LCS between the system and human summaries shows more similarity and therefore higher quality of the system summary.

\item \rg-W:
One problem with \rg-L is that all LCS with same lengths are rewarded equally. The LCS can be either related to a consecutive set of words or a long sequence with many gaps. While \rg-L treats all sequence matches equally, it makes sense that sequences with many gaps receive lower scores in comparison with consecutive matches. \rg-W considers an additional weighting function that awards consecutive matches more than non-consecutive ones.

\item \rg-S:
\rouge-S computes the skip-bigram co-occurrence statistics between the two summaries. It is similar to \rouge-2 except that it allows gaps between the bigrams by skipping middle tokens.

\item \rg-SU:
\rg-S does not give any credit to a system generated sentence if the sentence does not have any word pair co-occurring in the reference sentence. To solve this potential problem, \rg-SU was proposed which is an extension of \rg-S that also considers unigram matches between the two summaries.
\end{itemize}

\rg-L, \rg-W, \rg-S and \rg-SU were later extended to consider both the recall and precision. In calculating \rouge, stopword removal or stemming can also be considered, resulting in more variants. 

In the summarization literature, despite the large number of variants of \rouge, only one or very few of these variants are often chosen (arbitrarily) for evaluation of the quality of the summarization approaches. When \rouge was proposed, the original variants were only recall-oriented and hence the reported correlation results \cite{lin2004rouge}. The later extension of \rouge family by precision were only reflected in the later versions of the \rouge toolkit and additional evaluation of its effectiveness was not reported. Nevertheless, later published work in summarization adopted this toolkit for its ready implementation and relatively efficient performance. 

The original \rouge metrics show high correlations with human judgments of the quality of summaries on the DUC 2001-2003 benchmarks. However, these benchmarks consist of newswire data and are intrinsically very different than other summarization tasks such as summarization of scientific papers. We argue that \rouge is not the best metric for all summarization tasks and we propose an alternative metric for evaluation of scientific summarization. The proposed alternative metric shows much higher and more consistent correlations with manual judgments in comparison with the well-established \rouge.

\section{Summarization Evaluation by Relevance Analysis (\ser)}
\label{sec-method}
\textsc{Rouge} functions based on the assumption that in order for a summary to be of high quality, it has to share many words or phrases with a human gold summary. However, different terminology may be used to refer to the same concepts and thus relying only on lexical overlaps may underrate content quality scores. To overcome this problem, we propose an approach based on the premise that concepts take meanings from the context they are in, and that related concepts co-occur frequently. 

Our proposed metric is based on analysis of the content relevance between a system generated summary and the corresponding human written gold-standard summaries.
On high level, we indirectly evaluate the content relevance between the candidate summary and the human summary using information retrieval. To accomplish this, we use the summaries as search queries and compare the overlaps of the retrieved results. Larger number of overlaps, suggest that the candidate summary has higher content quality with respect to the gold-standard. This method, enables us to also reward for terms that are not lexically equivalent but semantically related. Our method is based on the well established linguistic premise that semantically related words occur in similar contexts \cite{turney2010frequency}. The context of the words can be considered as surrounding words, sentences in which they appear or the documents. For scientific summarization, we consider the context of the words as the scientific articles in which they appear. Thus, if two concepts appear in identical set of articles, they are semantically related. We consider the two summaries as similar if they refer to same set of articles even if the two summaries do not have high lexical overlaps. To capture if a summary relates to a article, we use information retrieval by considering the summaries as queries and the articles as documents and we rank the articles based on their relatedness to a given summary. For a given pair of system summary and the gold summary, similar rankings of the retrieved articles suggest that the summaries are semantically related, and thus the system summary is of higher quality.

Based on the domain of interest, we first construct an index from a set of articles in the same domain. Since TAC 2014 was focused on summarization in the biomedical domain, our index also comprises of biomedical articles. Given a candidate summary $C$ and a set of gold summaries $G_i$ ($i=1,...,M$; $M$ is the total number of human summaries), we submit the candidate summary and gold summaries to the search engine as queries and compare their ranked results. Let $I=\langle d_1,...,d_N \rangle$ be the entire index which comprises of $N$ total documents. 

 Let $R_C=\langle d_{\ell_1},...,d_{\ell_n} \rangle$ be the ranked list of retrieved documents for candidate summary $C$, and $R_{G_i}=\langle d_{\ell_1^{(i)}},...,d_{\ell_n^{(i)}} \rangle$ the ranked list of results for the gold summary $G_i$. These lists of results are based on a rank cut-off point $n$ that is a parameter of the system. We provide evaluation results on different choices of cut-off point $n$ in the Section \ref{sec:res} We consider the following two scores: (\textit{i}) simple intersection and (\textit{ii}) discounted intersection by rankings. The simple intersection just considers the overlaps of the results in the two ranked lists and ignores the rankings. The discounted ranked scores, on the other hand, penalizes ranking differences between the two result sets. As an example consider the following list of retrieved documents (denoted by $d_i$s) for a candidate and a gold summary as queries:

\textit{Results for candidate summary: $\langle d_1$, $d_2$, $d_3$, $d_4\rangle$}

\textit{Results for gold summary: $\langle d_3$, $d_2$, $d_1$, $d_4 \rangle$}

These two sets of results consist of identical documents but the ranking of the retrieved documents differ. Therefore, the simple intersection method assigns a score of 1.0 while in the discounted ranked score, the score will be less than 1.0 (due to ranking differences between the result lists).

We now define the metrics more precisely. Using the above notations, without loss of generality, we assume that $|R_C|\ge|R_{G_i}|$. \ser is defined as follows: 

\[
\textsc{Sera} = \frac{1}{M}\sum_{i=1}^M \frac{|R_C \cap R_{G_i}|}{|R_C|}
\]

To also account for the ranked position differences, we modify this score to discount rewards based on rank differences. That is, in ideal score, we want search results from candidate summary ($R_C$) to be the same as results for gold-standard summaries ($R_G$) and the rankings of the results also be the same. If the rankings differ, we discount the reward by log of the differences of the ranks. More specifically, the discounted score (\textsc{Sera-Dis}) is defined as:

\[
\footnotesize
\textsc{Sera-dis}=\frac{\sum\limits_{i=1}^{M}\big(\sum\limits_{j=1}^{|R_C|} \sum\limits_{k=1}^{|R_{G_i}|} 
\begin{cases}  (\frac{1}{\log(|j-k|+2)}) & \text{if } R_C^{(j)}=R_{G_i}^{(k)} \\
0 & \text{otherwise}
\end{cases}
 \big)}{M\times D_{\max}}
\]
\normalsize


where, as previously defined, $M$, $R_C$ and $R_{G_i}$ are total number of human gold summaries, result list for the candidate summary and result list for the human gold summary, respectively. In addition, $R_C^{(j)}$ shows the $j$th results in the ranked list $R_C$ and $D_{max}$ is the maximum attainable score used as the normalizing factor.  

We use elasticsearch\footnotemark[1], an open-source search engine, for indexing and querying the articles. For retrieval model, we use the Language Modeling retrieval model with Dirichlet smoothing \cite{zhai2001study}. Since TAC 2014 benchmark is on summarization of biomedical articles, the appropriate index would be the one constructed from articles in the same domain. Therefore, we use the open access subset of Pubmed\footnotemark[2] which consists of published articles in biomedical literature. 

\footnotetext[1]{\url{https://github.com/elastic/elasticsearch}}
\footnotetext[2]{PubMed is a comprehensive resource of articles and abstracts published in life sciences and biomedical literature \url{http://www.ncbi.nlm.nih.gov/pmc/}}

We also experiment with different query (re)formulation approaches. Query reformulation is a method in Information Retrieval that aims to refine the query for better retrieval of results. Query reformulation methods often consist of removing ineffective terms and expressions from the query (query reduction) or adding terms to the query that help the retrieval (query expansion). Query reduction is specially important when queries are verbose. Since we use the summaries as queries, the queries are usually long and therefore we consider query reductions.

In our experiments, the query reformulation is done by 3 different ways: (\textit{i}) Plain: The entire summary without stopwords and numeric values; (\textit{ii}) Noun Phrases (NP): We only keep the noun phrases as informative concepts in the summary and eliminate all other terms; and (\textit{iii}) Keywords (KW): We only keep the keywords and key phrases in the summary. For extracting the keywords and keyphrases (with length of up to 3 terms), we extract expressions whose \textit{idf}\footnotemark[1] values is higher than a predefined threshold that is set as a parameter. We set this threshold to the average \textit{idf} values of all terms except stopwords. \textit{idf} values are calculated on the same index that is used for the retrieval.

\footnotetext[1]{Inverted Document Frequency}

We hypothesize that using only informative concepts in the summary prevents query drift and leads to retrieval of more relevant documents. Noun phrases and keywords are two heuristics for identifying the informative concepts.

\section{Experimental setup}
\subsection{Data}
To the best of our knowledge, the only scientific summarization benchmark is from TAC 2014 summarization track. For evaluating the effectiveness of \rouge variants and our metric (\ser), we use this benchmark, which consists of 20 topics each with a biomedical journal article and 4 gold human written summaries. 

\subsection{Annotations}
In the TAC 2014 summarization track, \textsc{Rouge} was suggested as the evaluation metric for summarization and no human assessment was provided for the topics. Therefore, to study the effectiveness of the evaluation metrics, we use the semi-manual Pyramid evaluation framework \cite{nenkova2004pyramid,nenkova2007pyramid}. In the pyramid scoring, the content units in the gold human written summaries are organized in a pyramid. In this pyramid, the content units are organized in tiers and higher tiers of the pyramid indicate higher importance. The content quality of a given candidate summary is evaluated with respect to this pyramid.

To analyze the quality of the evaluation metrics, following the pyramid framework, we design an annotation scheme that is based on identification of important content units. Consider the following example:

\textit{\underline{Endogeneous small RNAs} (\underline{miRNA}) were genetically screened and studied to find the \underline{miRNAs} which are related to \underline{tumorigenesis}.}

In the above example, the underlined expressions are the content units that convey the main meaning of the text. We call these small units, nuggets which are phrases or concepts that are the main contributors to the content quality of the summary.

We asked two human annotators to review the gold summaries and extract content units in these summaries. The pyramid tiers represent the occurrences of nuggets across all the human written gold-standard summaries, and therefore the nuggets are weighted based on these tiers. The intuition is that, if a nugget occurs more frequently in the human summaries, it is a more important contributor (thus belongs to higher tier in the pyramid). Thus, if a candidate summary contains this nugget, it should be rewarded more. An example of the nuggets annotations in pyramid framework is shown in Table \ref{annexample}. In this example, the nugget ``\textit{cell mutation}'' belongs to the 4th tier and it suggests that the ``\textit{cell mutation}'' nugget is a very important representative of the content of the corresponding document.

Let $T_i$ define the tiers of the pyramid with $T_1$ being the bottom tier and $T_n$ the top tier. Let $N_i$ be the number of the nuggets in the candidate summary that appear in the tier $T_i$. Then the pyramid score $P$ of the candidate summary will be: 

$$P=\frac{1}{P_{\mathrm{max}}}\sum\limits_{i=1}^n i\times N_i$$

where $P_{\mathrm{max}}$ is the maximum attainable score used for normalizing the scores:

$$P_{\max}= \sum\limits_{i=j+1}^n i \times |T_i| + j \times (X - \sum\limits_{i=j+1}^n |T_i|) $$

where $X$ is the total number of nuggets in the summary and $j=\underset{i}{\max}\;\sum\limits_{t=i}^n {|T_t|} \ge X$.

We release the pyramid annotations of the TAC 2014 dataset through a public repository\footnotemark[2].

\footnotetext[2]{\url{https://github.com/acohan/TAC-pyramid-Annotations}}

\setlength{\dashlinedash}{3.1pt}

\begin{table}
\renewcommand{\arraystretch}{0.9}
\centering

\vspace{10pt}
\begin{tabular}{p{1cm}p{4cm}p{1cm}} \hline
id & nugget                               & Tier \\ \hline
$n_1$  & IDH1/2                               & 3                               \\ \hdashline
$n_2$  & isocitrate ~ ~ ~ ~~~~~~~ dehydrogenase 1 \& 2     & 2                               \\ \hdashline
$n_3$  & alpha ketoglutarate-dependent enzyme & 1                               \\ \hdashline
$n_4$  & TET2                                 & 1                               \\ \hdashline
$n_5$   & cell mutation                        & 4                               \\ \hdashline
$n_6$   & DNA methylation                      & 2   \\ \hline                        
\end{tabular}
\caption{Example of nugget annotation for Pyramid scores. The pyramid tier represents the number of occurrences of the nugget in all the human written gold summaries.}
\label{annexample}
\end{table}

\subsection{Summarization approaches}
\label{subsec:summ}
We study the effectiveness of \rouge and our proposed method (\ser) by analyzing the correlations with semi-manual human judgments.
Very few teams participated in TAC 2014 summarization track and the official results and the review paper of TAC 2014 systems were never published. Therefore, to evaluate the effectiveness of \rouge, we applied 9 well-known summarization approaches on the TAC 2014 scientific summarization dataset. Obtained \rouge and \ser results of each of these approaches are then correlated with semi-manual human judgments. In the following, we briefly describe each of these summarization approaches.

\begin{enumerate}[wide, labelwidth=!, labelindent=3pt]
\item LexRank \cite{erkan2004}: LexRank finds the most important (central) sentences in a document by using random walks in a graph constructed from the document sentences. In this graph, the sentences are nodes and the similarity between the sentences determines the edges. Sentences are ranked according to their importance. Importance is measured in terms of centrality of the sentence --- the total number of edges incident on the node (sentence) in the graph. The intuition behind LexRank is that a document can be summarized using the most central sentences in the document that capture its main aspects.

\item Latent Semantic Analysis (LSA) based summarization \cite{steinberger2004lsa}: In this summarization method, Singular Value Decomposition (SVD) \cite{deerwester1990indexing} is used for deriving latent semantic structure of the document. The document is divided into sentences and a term-sentence matrix $\mathbf{A}$ is constructed. The matrix $\mathbf{A}$ is then decomposed into a number of linearly-independent singular vectors which represent the latent concepts in the document. This method, intuitively, decomposes the document into several latent topics and then selects the most representative sentences for each of these topics as the summary of the document.


\item Maximal Marginal Relevance (MMR) \cite{carbonell1998use}:
Maximal Marginal Relevance (MMR) is a greedy strategy for selecting sentences for the summary. Sentences are added iteratively to the summary based on their relatedness to the document as well as their novelty with respect to the current summary.

\item Citation based summarization \cite{qazvinian2013generating}: In this method, citations are used for summarizing an article. Using the LexRank algorithm on the citation network of the article, top sentences are selected for the final summary.

\item Using frequency of the words \cite{luhn1958automatic}: In this method, which is one the earliest works in text summarization, raw word frequencies are used to estimate the saliency of sentences in the document. The most salient sentences are chosen for the final summary.

\item SumBasic \cite{vanderwende2007beyond}:
SumBasic is an approach that weights sentences based on the distribution of words that is derived from the document. Sentence selection is applied iteratively by selecting words with highest probability and then finding the highest scoring sentence that contains that word. The word weights are updated after each iteration to prevent selection of similar sentences.

\item Summarization using citation-context and discourse structure \cite{cohan2015scientific}: In this method, the set of citations to the article are used to find the article sentences that directly reflect those citations (citation-contexts). In addition, the scientific discourse of the article is utilized to capture different aspects of the article. The scientific discourse usually follows a structure in which the authors first describe their hypothesis, then the methods, experiment, results and implications. Sentence selection is based on finding the most important sentences in each of the discourse facets of the document using the MMR heuristic.

\item KL Divergence \cite{haghighi2009exploring}
In this method, the document unigram distribution $P$ and the summary unigram distributation $Q$ are considered; the goal is to find a summary whose distribution is very close to the document distribution. The difference of the distributions is captured by the Kullback-Lieber (KL) divergence, denoted by $\mathrm{KL}(P||Q)$.

\item Summarization based on Topic Models \cite{haghighi2009exploring}: Instead of using unigram distributions for modeling the content distribution of the document and the summary, this method models the document content using an LDA based topic model \cite{blei2003latent}. It then uses the KL divergence between the document and the summary content models for selecting sentences for the summary.

\end{enumerate}
\section{Results and Discussion}
\label{sec:res}

\begin{table}[t]
\renewcommand{\arraystretch}{0.9}
\small
\centering
\begin{tabular}{l|ccc}
               & \multicolumn{3}{c}{Pyramid}         \\ \hline
Metric         & Pearson($r$) & Spearman($\rho$) & Kendall($\tau$) \\ \hline
\rg-1-F      & 0.454   & 0.174    & 0.138   \\
\rg-1-P      & 0.257   & 0.116    & 0       \\
\rg-1-R      & 0.513   & 0.229    & 0.138       \\
\rg-2-F      & 0.816   & 0.696    & 0.552   \\
\rg-2-P      & 0.824   & 0.841    & 0.69    \\
\rg-2-R      & 0.803   & 0.696    & 0.552   \\
\textbf{\rg-3-F}      & \textbf{0.878}   & 0.841    & 0.69    \\
\rg-3-P      & 0.875   & 0.725    & 0.552   \\
\rg-3-R      & 0.875   & 0.841    & 0.69    \\
\rg-L-F      & 0.454   & 0.261    & 0.276   \\
\rg-L-P      & 0.262   & 0.29     & 0.138   \\
\rg-L-R      & 0.52    & 0.261    & 0.276   \\
\rg-S-F     & 0.603   & 0.406    & 0.414   \\
\rg-S-P     & 0.344   & 0.174    & 0.138   \\
\rg-S-R     & 0.664   & 0.406    & 0.414   \\
\rg-SU-F    & 0.601   & 0.493    & 0.462   \\
\rg-SU-P    & 0.338   & 0.174    & 0.138   \\
\rg-SU-R    & 0.662   & 0.406    & 0.414   \\
\rg-W-1.2-F  & 0.607   & 0.493    & 0.414   \\
\rg-W-1.2-P  & 0.418   & 0.377    & 0.276   \\
\rg-W-1.2-R  & 0.626   & 0.667    & 0.552   \\ \hdashline
\sr-5       & 0.823   & 0.941    & 0.857   \\
\sr-10       & 0.788   & 0.647    & 0.429   \\
\sr-KW-5    & 0.848   & 0.765    & 0.571   \\
\sr-KW-10    & 0.641   & 0.618    & 0.486   \\
\textbf{\sr-NP-5}    & 0.859   & \textbf{1.0}       & \textbf{1.0}       \\
\sr-NP-10    & 0.806   & 0.941    & 0.857   \\
\sr-DIS-5    & 0.631   & 0.824    & 0.714   \\
\sr-DIS-10    & 0.687   & 0.824    & 0.714   \\
\sr-DIS-KW-5 & 0.838   & 0.941    & 0.857   \\
\sr-DIS-KW-10 & 0.766   & 0.712    & 0.729   \\
\sr-DIS-NP-5 & 0.834   & 0.941    & 0.857   \\
\sr-DIS-NP-10 & 0.86    & 0.941    & 0.857  \\ \hline
\end{tabular}
\caption{\footnotesize Correlation between variants of \textsc{Rouge} and \ser, with human pyramid scores. All variants of \rouge are displayed. $F$: F-Score; $R$: Recall; $P$: Precision; DIS: Discounted variant of \ser; KW: using Keyword query reformulation; NP: Using noun phrases for query reformulation. The numbers in front of the \ser metrics indicate the rank cut-off point.}
\label{res-cor}
\end{table}

We calculated all variants of \textsc{Rouge} scores, our proposed metric, \ser, and the Pyramid score on the generated summaries from the summarizers described in Section \ref{subsec:summ}. We do not report the \rouge, \ser or pyramid scores of individual systems as it is not the focus of this study. Our aim is to analyze the effectiveness of the evaluation metrics, not the summarization approaches. Therefore, we consider the correlations of the automatic evaluation metrics with the manual Pyramid scores to evaluate their effectiveness; the metrics that show higher correlations with manual judgments are more effective.

Table \ref{res-cor} shows the Pearson, Spearman and Kendall correlation of \textsc{Rouge} and \ser, with pyramid scores. Both \rouge and \ser are calculated with stopwords removed and with stemming. Our experiments with inclusion of stopwords and without stemming showed similar results and thus, we do not include those to avoid redundancy. 

\subsection{\ser}
The results of our proposed method (\ser) are shown in the bottom part of Table \ref{res-cor}. In general, \ser shows better correlation with pyramid scores in comparison with \rouge. We observe that the Pearson correlation of \ser with cut-off point of 5 (shown by \textit{\ser}-5) is 0.823 which is higher than most of the \rouge variants. Similarly, the Spearman and Kendall correlations of the \ser evaluation score is 0.941 and 0.857 respectively, which are higher than all \rouge correlation values. 
 This shows the effectiveness of the simple variant of our proposed summarization evaluation metric.

Table \ref{res-cor} also shows the results of other \ser variants including discounting and query reformulation methods. Some of these variants are the result of applying query reformulation in the process of document retrieval which are described in section \ref{sec-method} As illustrated, the Noun Phrases (NP) query reformulation at cut-off point of 5 (shown as \textsc{\sr-np-5}) achieves the highest correlations among all the \ser variants ($r\;$=$\;0.859$, $\rho\;$=$\;\tau\;$=$\;1.0$). In the case of Keywords (KW) query reformulation, without using discounting, we can see that there is no positive gain in correlation. However, keywords when applied on the discounted variant of \ser, result in higher correlations.

Discounting has more positive effect when applied on query reformulation-based \ser than on the simple variant of \ser. In the case of discounting and NP query reformulation (\textsc{\sr-dis-np}), we observe higher correlations in comparison with simple \ser. Similarly, in the case of Keywords (KW), positive correlation gain is obtained in most of correlation coefficients. NP without discounting and at cut-off point of 5 (\textsc{\sr-np-5}) shows the highest non-parametric correlation. In addition, the discounted NP at cut-off point of 10 (\textsc{\sr-np-dis-10}) shows the highest parametric correlations.

In general, using NP and KW as heuristics for finding the informative concepts in the summary effectively increases the correlations with the manual scores. Selecting informative terms from long queries results in more relevant documents and prevents query drift. Therefore, the overall similarity between the two summaries (candidate and the human written gold summary) is better captured.

\begin{table}[]
\small
\setlength{\tabcolsep}{6pt}
\renewcommand{\arraystretch}{0.9}
\centering
\begin{tabular}{l|ccc|ccc}
               & \multicolumn{3}{c}{\rg-2-F}       & \multicolumn{3}{c}{\rg-3-F}              \\ \hline
Metric         & $r$ & $\rho$ & $\tau$ & $r$ & $\rho$ & $\tau$ \\ \hline
\sr-5       & .408     & .522     & .414     & .540      & .725     & .552     \\
\sr-10       & .447     & .406     & .276     & 0.6       & .667     & .414     \\
\textbf{\sr-KW-5}    & .867     & .754     & .690      & .770      & .899     & .828     \\
\sr-KW-10    & .574     & .174     & .138     & .343     & .029     & 0         \\
\sr-NP-5    & .588     & .696     & .552     & .720      & .841     & .690      \\
\sr-NP-10    & .416     & .522     & .414     & .609     & .725     & .552     \\
\sr-DIS-5    & .154     & .464     & .276     & .396     & .667     & .414     \\
\sr-DIS-10    & .280      & .464     & .276     & .502     & .667     & .414     \\
\textbf{\sr-DIS-KW-5} & .891     & .812     & .690      & .842     & .899     & .828     \\
\sr-DIS-KW-10 & .751     & .696     & .552     & .650      & .551     & .414     \\
\sr-DIS-NP-5 & .584     & .522     & .414     & .744     & .725     & .552     \\
\sr-DIS-NP-10 & .583     & .522     & .414     & .763     & .725     & .552      \\ \hline
\end{tabular}
\caption{Correlation between \ser and \rouge scores. NP: Query reformulation with Noun Phrases; KW: Query reformulation with Keywords; DIS: Discounted variant of \ser; The numbers in front of the \ser metrics indicate the rank cut-off point.}
\label{res-rouge-relation}
\end{table}

\subsection{\rouge}
Another important observation is regarding the effectiveness of \textsc{Rouge} scores (top part of Table \ref{res-cor}). Interestingly, we observe that many variants of \textsc{Rouge} scores do not have high correlations with human pyramid scores. The lowest F-score correlations are for \textsc{Rouge-1} and \textsc{Rouge-L} (with $r$=0.454). Weak correlation of \textsc{Rouge-1} shows that matching unigrams between the candidate summary and gold summaries is not accurate in quantifying the quality of the summary. On higher order n-grams, however, we can see that \textsc{Rouge} correlates better with pyramid. In fact, the highest overall $r$ is obtained by \textsc{Rouge-3}. \textsc{Rouge-L} and its weighted version \textsc{Rouge-W}, both have weak correlations with pyramid. Skip-bigrams (\textsc{Rouge-S}) and its combination with unigrams (\textsc{Rouge-SU}) also show sub-optimal correlations. Note that $\rho$ and $\tau$ correlations are more reliable in our setup due to the small sample size.

These results confirm our initial hypothesis that \textsc{Rouge} is not accurate estimator of the quality of the summary in scientific summarization. We attribute this to the differences of scientific summarization with general domain summaries. When humans summarize a relatively long research paper, they might use different terminology and paraphrasing. Therefore, \rouge which only relies on term matching between a candidate and a gold summary, is not accurate in quantifying the quality of the candidate summary.

\subsection{Correlation of \textsc{\ser} with \rouge}
Table \ref{res-rouge-relation} shows correlations of our metric \ser with \rg-2 and \rg-3, which are the highest correlated \rouge variants with pyramid. We can see that in general, the correlation is not strong. Keyword based reduction variants are the only variants for which the correlation with \rouge is high. Looking at the correlations of KW variants of \ser with pyramid (Table \ref{res-cor}, bottom part), we observe that these variants are also highly correlated with manual evaluation. 

\begin{figure}
\centering
\includegraphics[width=\linewidth,height=5cm]{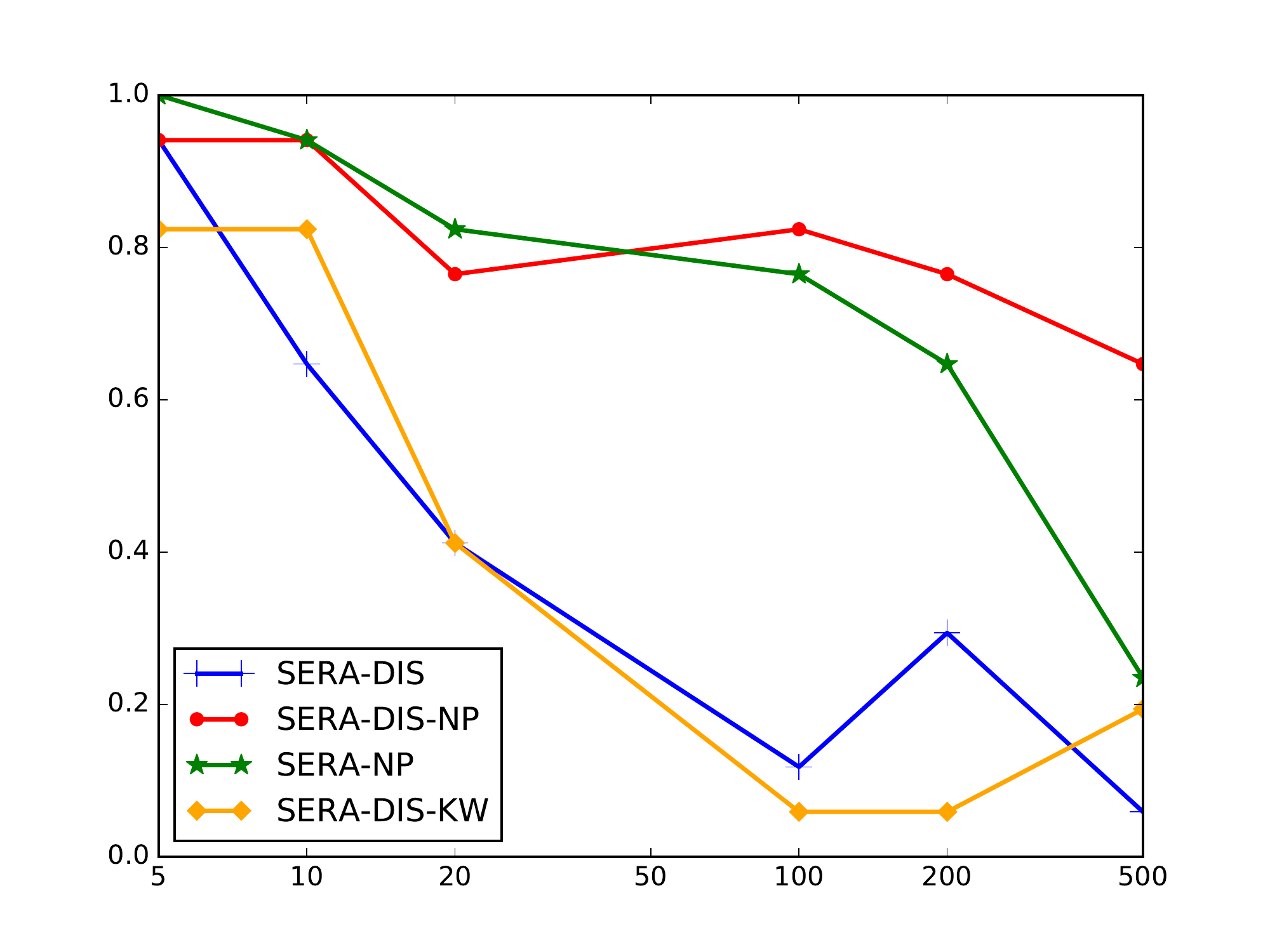}
\caption{$\rho$ correlation of \ser with pyramid based on different cut-off points. The x-axis shows the cut-off point parameter. DIS: Discounted variant of \ser; NP: Query reformulation with Noun Phrases; KW: Query reformulation with Keywords.}
\label{fig-cor}
\end{figure}

\subsection{Effect of the rank cut-off point}
Finally, Figure \ref{fig-cor} shows $\rho$ correlation of different variants of \ser with pyramid based on selection of different cut-off points ($r$ and $\tau$ correlations result in very similar graphs). When the cut-off point increases, more documents are retrieved for the candidate and the gold summaries, and therefore the final \ser score is more fine-grained. A general observation is that as the search cut-off point increases, the correlation with pyramid scores decreases. This is because when the retrieved result list becomes larger, the probability of including less related documents increases which negatively affects correct estimation of the similarity of the candidate and gold summaries. The most accurate estimations are for metrics with cut-off points of 5 and 10 which are included in the reported results of all variants in Table \ref{res-cor}. 

\section{Related work}

\textsc{Rouge} \cite{lin2004rouge} assesses the content quality of a candidate summary with respect to a set of human gold summaries based on their lexical overlaps. \textsc{Rouge} consists of several variants. Since its introduction, \textsc{Rouge} has been one of the most widely reported metrics in the summarization literature, and its high adoption has been due to its high correlation with human assessment scores in DUC datasets \cite{lin2004rouge}. However, later research has casted doubts about the accuracy of \rouge against manual evaluations. \newcite{conroy2008mind} analyzed DUC 2005 to 2007 data and showed that while some systems achieve high \rouge scores with respect to human summaries, the linguistic and responsiveness scores of those systems do not correspond to the high \rouge scores.  

We studied the effectiveness of \rouge through correlation analysis with manual scores. Besides correlation with human assessment scores, other approaches have been explored for analyzing the effectiveness of summarization evaluation. \newcite{Rankel:2011} studied the extent to which a metric can distinguish between the human and system generated summaries. They also proposed the use of paired two-sample t-tests and the Wilcoxon signed-rank test as an alternative to \rouge in evaluating several summarizers. Similarly, \newcite{owczarzak2012assessment} proposed the use of multiple binary significance tests between the system summaries for ranking the best summarizers.

Since introduction of \rouge, there have been other efforts for improving automatic summarization evaluation. \newcite{hovy2006automated} proposed an approach based on comparison of so called Basic Elements (BE) between the candidate and reference summaries. BEs were extracted based on syntactic structure of the sentence. The work by \newcite{conroy2011nouveau} was another attempt for improving \textsc{Rouge} for update summarization which combined two different \textsc{Rouge} variants and showed higher correlations with manual judgments for TAC 2008 update summaries.

 Apart from the content, other aspects of summarization such as linguistic quality have been also studied. \newcite{pitler2010automatic} evaluated a set of models based on syntactic features, language models and entity coherences for assessing the linguistic quality of the summaries. Machine translation evaluation metrics such as \textsc{blue} have also been compared and contrasted against \textsc{Rouge} \cite{graham:2015:EMNLP}. Despite these works, when gold-standard summaries are available, \textsc{Rouge} is still the most common evaluation metric that is used in the summarization published research. Apart from \rg's initial good results on the newswire data, the availability of the software and its efficient performance have further contributed to its popularity.

\section{Conclusions}

We provided an analysis of existing evaluation metrics for scientific summarization with evaluation of all variants of \rouge. We showed that \rouge may not be the best metric for summarization evaluation; especially in summaries with high terminology variations and paraphrasing (e.g. scientific summaries). Furthermore, we showed that different variants of \rouge result in different correlation values with human judgments, indicating that not all \rouge scores are equally effective. Among all variants of \rouge, \textsc{Rouge-2} and \textsc{Rouge-3} are better correlated with manual judgments in the context of scientific summarization. We furthermore proposed an alternative and more effective approach for scientific summarization evaluation (Summarization Evaluation by Relevance Analysis - \ser). Results revealed that in general, the proposed evaluation metric achieves higher correlations with semi-manual pyramid evaluation scores in comparison with \rouge.

Our analysis on the effectiveness of evaluation measures for scientific summaries was performed using correlations with manual judgments. An alternative approach to follow would be to use statistical significance testing on the ability of the metrics to distinguish between the summarizers (similar to \newcite{Rankel:2011}). We studied the effectiveness of existing summarization evaluation metrics in the scientific text genre and proposed an alternative superior metric. Another extension of this work would be to evaluate automatic summarization evaluation in other genres of text (such as social media). Our proposed method only evaluates the content quality of the summary. Similar to most of existing summarization evaluation metrics, other qualities such as linguistic cohesion, coherence and readability are not captured by this method. Developing metrics that also incorporate these qualities is yet another future direction to follow.

\section*{Acknowledgments}
We would like to thank all three anonymous reviewers for their feedback and comments, and Maryam Iranmanesh for helping in annotation. This work was partially supported by National Science Foundation (NSF) through grant CNS-1204347.


\section{Bibliographical References}
\label{main:ref}

\bibliographystyle{lrec2016}


\end{document}